\documentclass[10pt,conference]{IEEEtran}
\IEEEoverridecommandlockouts

\usepackage{cite}
\usepackage{booktabs}

\usepackage{amsmath,amssymb,amsfonts,amsthm}

\usepackage[ruled,linesnumbered]{algorithm2e}
\SetKwComment{Comment}{/* }{ */}

\usepackage{fancyhdr}
\theoremstyle{definition}

\usepackage{caption}
\usepackage{graphicx}
\usepackage{textcomp}
\usepackage{xcolor}
\usepackage{tikz}

\newcommand*\circled[1]{\tikz[baseline=(char.base)]{
\node[shape=circle,draw,inner sep=1pt] (char) {#1};}}
\def\BibTeX{{\rm B\kern-.05em{\sc i\kern-.025em b}\kern-.08em
T\kern-.1667em\lower.7ex\hbox{E}\kern-.125emX}}

\usepackage{hyperref}
\hypersetup{
    colorlinks=true,
    linkcolor=blue,
    filecolor=magenta,      
    urlcolor=cyan,
    }
\usepackage{multicol}
\usepackage{multirow}

\usepackage{listings}

\lstdefinestyle{mystyle}{
    backgroundcolor=\color{backcolour},   
    commentstyle=\color{codegreen},
    keywordstyle=\color{magenta},
    numberstyle=\tiny\color{codegray},
    stringstyle=\color{codepurple},
    basicstyle=\ttfamily\footnotesize,
    breakatwhitespace=false,         
    breaklines=true,                 
    captionpos=b,                    
    keepspaces=true,                 
    numbers=left,                    
    numbersep=5pt,                  
    showspaces=false,                
    showstringspaces=false,
    showtabs=false,                  
    tabsize=2
}
\usepackage{enumitem}
\setlist[itemize]{align=parleft,left=0pt..1em}
\usepackage[normalem]{ulem}
\usepackage{framed}
\definecolor{shadecolor}{gray}{0.9}

\begin{document}
%-------------------------------------------------------------------------------

\title{Understanding and Mitigating Network Latency Effect on Teleoperated-Robot with Extended Reality}

% If your conference documentclass or package defines these macros,
% change these macros to different names.
\newcommand*{\affaddr}[1]{#1} % No op here. Customize it for different styles.
\newcommand*{\affmark}[1][*]{\textsuperscript{#1}}
\newcommand*{\email}[1]{\texttt{#1}}

\author{%
Ziliang Zhang, Cong Liu, Hyoseung Kim\\
\affaddr{University of California, Riverside}\\
\email{\{zzhan357, congl, hyoseung\}@ucr.edu}%
}

\maketitle

\begin{abstract}

Robot teleoperation with extended reality (XR teleoperation) enables intuitive interaction by allowing remote robots to mimic user motions with real-time 3D feedback. However, existing systems face significant motion-to-motion (M2M) latency--the delay between the user's latest motion and the corresponding robot feedback--leading to high teleoperation error and mission completion time. This issue stems from the system's exclusive reliance on network communication, making it highly vulnerable to network degradation.
  
To address these challenges, we introduce TeleXR, the first end-to-end, fully open-sourced XR teleoperation framework that decouples robot control and XR visualization from network dependencies. TeleXR leverages local sensing data to reconstruct delayed or missing information of the counterpart, thereby significantly reducing network-induced issues. This approach allows both the XR and robot to run concurrently with network transmission while maintaining high robot planning accuracy. TeleXR also features contention-aware scheduling to mitigate GPU contention and bandwidth-adaptive point cloud scaling to cope with limited bandwidth.

\end{abstract}

\section{Introduction}

Robot teleoperation via Extended Reality (XR teleoperation) has recently gained significant attention for its intuitive user-robot interaction over long distances. It emerges as a transformative technology in industrial automation, surgical robotics, autonomous robot training, %\Cong{more citations, particularly within 2023-2025}
and humanoid robotics~\cite{naceri2019towards,xu2022design,davinci,zinchenko2021autonomous,black2024human,cheng2024open,iyer2024open,he2024omnih2o,hirschmanner2019virtual}. In XR teleoperation, an XR device employs multiple sensors to capture user motion and produce user poses that are transmitted to the remote robot. Guided by these poses, the robot imitates the user motion and returns its own pose and surrounding environment to the XR device. The XR device then generates a 3D frame using the latest user pose along with the received robot pose and surroundings, and displays it on a head-mounted display (HMD). By referencing both user and robot poses in the frame, the user executes subsequent motions, repeating the process until the target mission is completed.

Prior research has identified a significant delay between user motion and robot feedback in XR teleoperation, primarily caused by network latency and the extended control-feedback pipeline. 
%
% \Cong{Be consistent everywhere. You have three terms: Motion-to-motion, motion-to-motion, Motion-to-Motion.} \Johnson{Changed every occurance to motion-to-motion (M2M) latency}
Existing systems quantify this delay using the motion-to-motion (M2M) latency metric, which measures the time to display the robot motion triggered by the latest user motion sensed via XR camera~\cite{su2023latency,naceri2019towards}. Consequently, users experience a pronounced discrepancy between their own pose and the robot pose in the displayed 3D frame, referred to as teleoperation error. Continuing motion despite this error makes the robot deviate from the user's intention, forcing the user to postpone subsequent motions. Moreover, prolonged network latency and hardware constraints lead to longer M2M latency, while network fluctuations and bandwidth limitations further amplify the teleoperation error.

This work identifies the fundamental cause of these issues as the strong dependency on network communication in XR teleoperation. When generating each frame, the robot depends on XR–to-robot communication to receive the user pose for control, while XR relies on robot–to-XR communication to obtain the robot pose and surrounding information for visualization. This dependency makes XR teleoperation vulnerable to latency, fluctuations, and packet drops under real-world conditions. 
% Based on this insight, we for the first time propose a paradigm that reduces network dependency by enabling each side to locally reconstruct the other’s information. This paradigm enables overlapping execution of XR visualization and network transmission, while keeping robot control accuracy with locally generated waypoints, making continuous teleoperation possible.

\section{Challenges}
\label{sec:Motivation}

\subsection{Network and Hardware Challenges}\label{subsec:Network_and_Hardware_Challenges}

Building an XR teleoperation framework with low M2M latency, controllable teleoperation error, and short completion time is undoubtedly challenging. To uncover these challenges, we built a test system using a Northstar Next XR Headset\cite{northstar_next} and a Kinova Gen3~\cite{kinova} robot manipulator, each located in separate rooms. Both devices connect to separate routers with 5GHz 802.11ac Wireless Local Area Networks (WLAN) and communicate through a cloud VPN server in the same region to introduce realistic network latency. We evaluated the system under four configurations: In \circled{1}, both devices operate on PCs and adopt a VR gallery~\cite{vrgallery} scene when rendering the output 3D frame. In \circled{2}, XR hardware is switched to an NVIDIA Jetson Xavier~\cite{xavier} and robot host to an NVIDIA Orin Nano~\cite{nano} for limited computation power. In \circled{3}, PC hardware is used with a higher rendering demand Sponza~\cite{sponza} scene. In \circled{4}, the user operates in a dimly lit room with increased camera noise that prolongs XR tracking time.\label{subsec:mot_config}

\begin{figure}[t]
\includegraphics[width=\linewidth]{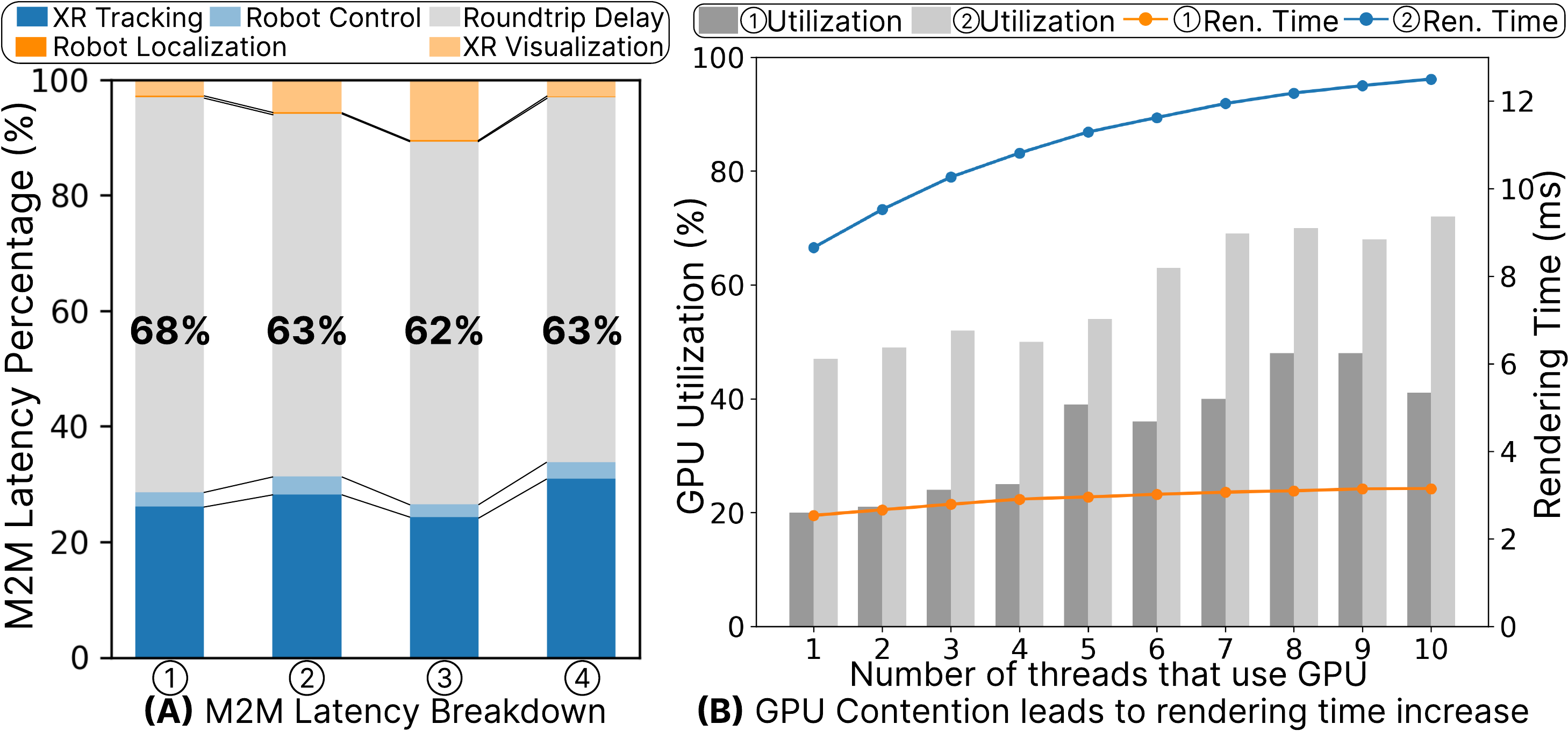}
\caption{Network and hardware effect. Circled numbers indicate configurations explained in \S\ref{subsec:mot_config}.}
\label{fig:m2m_affect}
\end{figure}

\textbf{Network Latency.} We decompose the M2M latency by profiling the round-trip network delay and the end-to-end latency across the four modules, each running as a separate process. Fig.~\ref{fig:m2m_affect}(A) shows that network delay is the dominant factor in M2M latency, regardless of variations in hardware, virtual scene, or operating environment.
Such network delay will become more significant when the XR and robot are geographically far apart. %Large network delay will directly increase M2M latency due to the network dependency in both robot control and XR visualization. 
In the test system, robot control cannot make subsequent commands before the arrival of user hand pose and fist gesture, while XR visualization cannot update the displayed robot until the arrival of the remote robot pose and gripper intensity.

\textbf{Hardware Contention.} State-of-the-art XR and robotic systems are deployed on embedded devices with limited computational resources, leaving them prone to performance degradation under heavy workloads. In current XR teleoperation systems, multiple functionalities, such as eye tracking, hand tracking, and pose estimation, rely heavily on GPU computation and are designed to run within the same process boundary. To quantify the impact of these hardware constraints, we profile execution time of the rendering functionality within XR visualization as rendering time. In our test system, rendering is implemented as a dedicated thread using the Vulkan 3D graphics API~\cite{sellers2016vulkan} to generate a 3D frame. We simulate increased workload by adding additional threads that utilize the same GPU stream for fixed matrix multiplications under configurations \circled{1} and \circled{2}, as the same computation widely exists in other XR visualization functionalities including reprojection, pose prediction, and background rendering~\cite{atw,taud2017multilayer,illixr,zhang2024boxr}. Fig.~\ref{fig:m2m_affect}(B) shows that as the number of threads increases, rendering time increases whereas GPU utilization may even decrease. This degradation occurs because multiple threads compete for the same GPU stream, leading to blocking of the original rendering thread. Noticeably, on configuration \circled{2} for embedded devices, such blocking causes rendering time to increase by up to 45\%, resulting in a noticeable drop in frame rate. 

% \subsection{Network Dynamics on Teleoperation Error}
\subsection{Network and Bandwidth Dynamics}\label{subsec:Network_and_Bandwidth Dynamics}
% \section{System Dynamics}

Apart from the fundamental challenges due to network and hardware, the runtime dynamics from network communication pose great threats to the XR teleoperation framework. To simulate the network dynamics, we change the network connection from the one used previously in \S\ref{subsec:Network_and_Hardware_Challenges} to 802.11b 2.4GHz WLAN and then switch to cellular network running on 4G LTE and 5G, while keeping the framework configuration the same as \circled{1}. 

\begin{figure}[t]
\includegraphics[width=\linewidth]{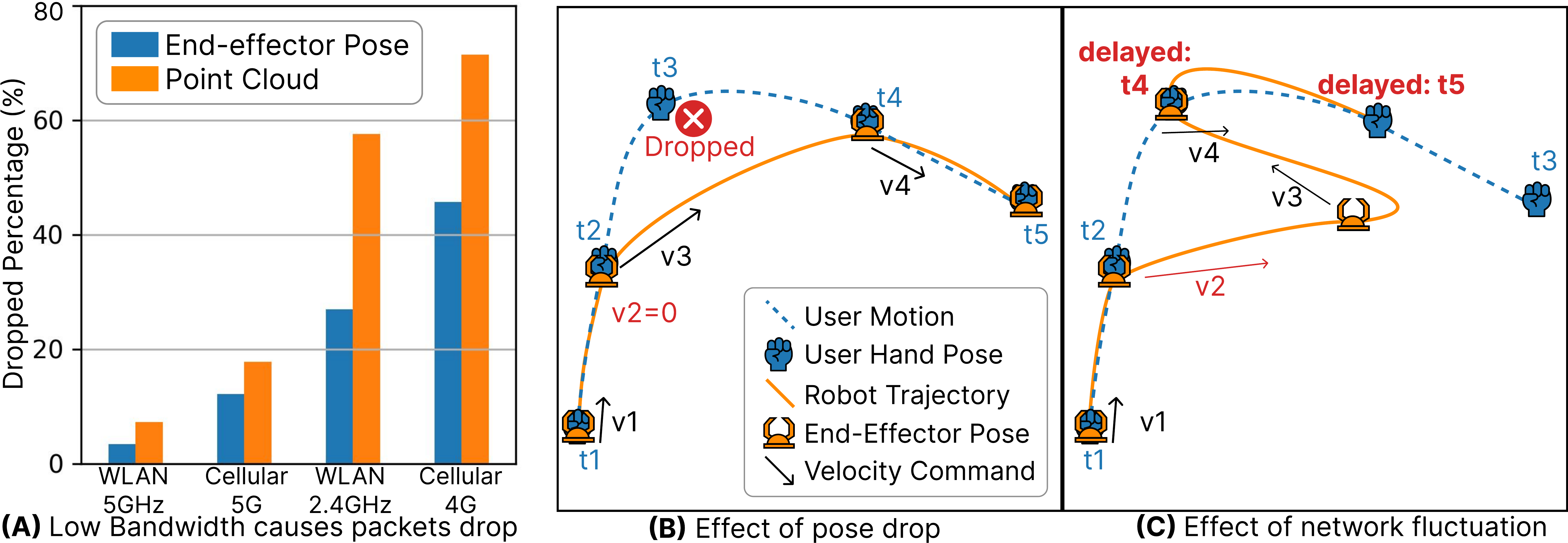}
\caption{Runtime dynamics and effect.}
\label{fig:tele_err_affect}
\end{figure}

\textbf{Packet Drops}. One immediate observation is the packet drops as we switch to communication with much limited bandwidth. Point cloud data, which ranges from 1.5 to 3 MB per period, is substantially larger than the other control data that typically measures only a few tens of bytes. We profile the received end-effector poses, gripper intensity, and point cloud on the XR side under four different network settings and compare them against the transmitted ones from robot side to calculate the drop percentage. Fig.~\ref{fig:tele_err_affect}(A) shows that as bandwidth decreases from 5GHz WLAN to Cellular 4G, all of the end-effector pose, gripper intensity, and point cloud experience more drops during teleoperation. The effect of these drops is illustrated in Fig.~\ref{fig:tele_err_affect}(B). In this example, five hand poses, sent sequentially from time $t_1$ to $t_5$, are intended to control the robot, but the $t_3$ pose is not transmitted to robot due to bandwidth limitation. As a result, when the robot reaches $t_2$ pose, it does not receive any new pose data and stops until $t_4$. Instead of calculating speed and direction to reach the missing $t_3$ pose, the robot generates command to move to $t_4$ pose. This abrupt change causes a teleoperation error and leads to an irrecoverable robot trajectory deviation.

\textbf{Network Fluctuation.} Since the XR device and the robot communicate wirelessly through a VPN server in our setup, network fluctuations occur during runtime because of random congestion and signal interference. These fluctuations can cause user poses for controlling the robot to arrive out of order. This effect is illustrated in Fig.~\ref{fig:tele_err_affect}(C), where the same user poses are transmitted, but the last pose arrives earlier than the preceding two. At time $t_2$, the robot receives the incorrect $t_3$ pose and moves toward it without waiting. When the delayed poses subsequently arrive at time $t_4$ and $t_5$, the robot moves toward them before reaching $t_3$ pose, resulting in a trajectory that deviates significantly from the intended motion. Although one may consider using timestamps to enforce a strict sequential order, this approach increases the robot's waiting time, forcing the user to pause longer for the robot to catch up and ultimately extending the completion time.

\begin{figure}[t]
\includegraphics[width=\linewidth]{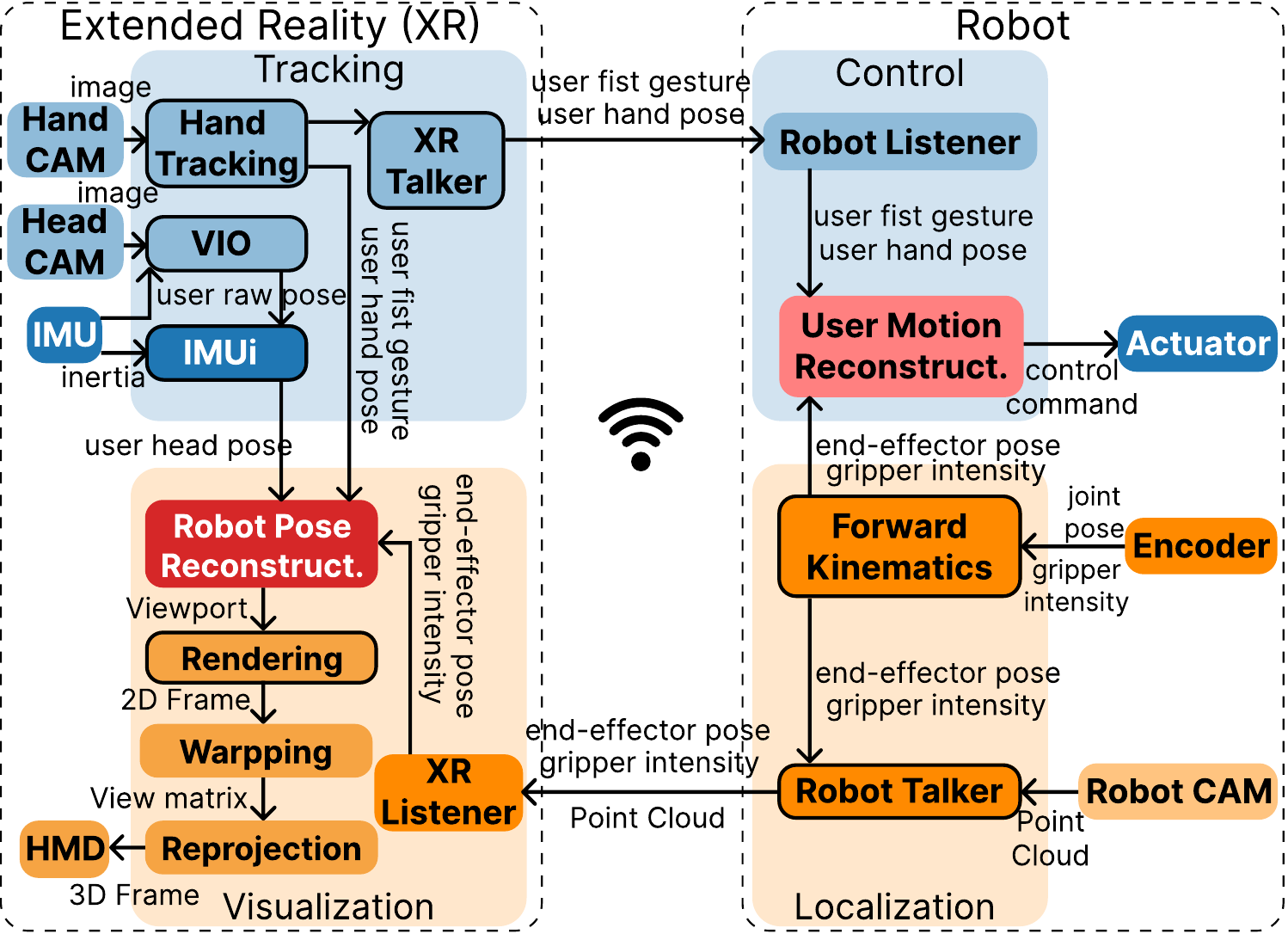}
\caption{TeleXR Framework Details.}
\label{fig:system_overview}
\end{figure}

\section{TeleXR Framework}

% \subsection{\Cong{How about using this title: The Dual-Reconstruction Software Architecture of TeleXR. If using this title, making sure explaining dual-reconstruction aspect clear enough and why it is a good idea.} Software Architecture}\label{subsec:Software_Architecture}

\subsection{Dual-Reconstruction Architecture}\label{subsec:Software_Architecture}

As illustrated in \S\ref{sec:Motivation}, existing XR teleoperation frameworks suffer from intrinsically high M2M latency and significant teleoperation error, causing robot trajectory to deviate from the user's intended motion. 
This is primarily due to their exclusive reliance on network communication to transmit user information for robot control and robot information for XR visualization. Any delay, jitter, or packet loss may lead to cascading performance degradation in these tightly coupled operations. 
We observe that since both XR device and remote robot possess sensing capability and operate in a shared coordinate system, delayed or missing information can be locally reconstructed by extrapolating the last received information according to the most recent local sensing data, decoupling the robot control and XR visualization from network dependency. Because the reconstruction occurs on both XR and robot sides, we introduce this paradigm as \textit{Dual-Reconstruction}.
%This dual-reconstruction paradigm decouples the robot control and XR visualization from network dependency, overlapping execution of XR visualization and network transmission to shorten M2M latency while keeping robot control accuracy.

% \Cong{check the accuracy and any overclaiming of the following paragraph. You want to highlight sth that convince readers this architecture is fundamentally new, otherwise it will read just like a typical overview at the beginning of the method section.}

TeleXR is the first end-to-end XR teleoperation framework that fundamentally redefines this architecture. Unlike previous systems that merely wait for delayed network responses, TeleXR adopts a novel dual-reconstruction paradigm that simultaneously decouples both control and feedback pipelines from network dependencies. 
%, enabling localized extrapolation of control and feedback signals at both ends. 
Diverging from the established systems, TeleXR incorporates the reconstructed information into the pipeline executions and constantly updates the reconstruction model based on newly received information.
This approach mitigates network degradation by allowing both the XR and robotic subsystems to continuously reconstruct each other’s state. It ensures that the robot proactively aligns with the user’s intended motion, significantly reducing the impact of communication delays and improving mission completion time with reduced user pauses. 
%{\red As illustrated by the Robot' motion from Fig.\ref{fig:system_overview}(A), XR locally reconstructs the robot motion prior to remote pose arrival, making robot motion closely align with the user motion.}
Fig.\ref{fig:system_overview}(A) gives an example where frame duration, robot movement, and one-way network communication all take one time unit. Although the actual robot pose at time 1 ($R_1$) is not immediately available to the XR device due to network delay, TeleXR leverages the user pose generated at time 1 ($U_1$) to reconstruct the expected robot pose at time 1 ($\hat{R}_1$) and produces frame 1 that displays both $U_1$ and $\hat{R}_1$. Simultaneously, the remote robot receives $U_0$ and reconstructs any missing or delayed user poses based on its current pose and speed.

Building on the dual-reconstruction paradigm, TeleXR implements two key optimizations to further enhance system performance: contention-aware scheduling and bandwidth-adaptive point cloud scaling. 
The contention-aware scheduling mitigates GPU contention by synchronizing functionalities execution within the correct pipeline, effectively regulating M2M latency. Bandwidth-adaptive point cloud scaling selectively omits the non-edge 3D points of the objects to reduce the transmission payload size during a limited bandwidth condition. 
Together, the proposed architecture facilitates stable, high-quality teleoperation in practical deployment scenarios, marking a paradigm shift from prior systems that are inherently susceptible to network variance.

\textbf{Detailed System Design.} We adopt a publisher-subscriber model for both the XR and robot to accommodate various functionalities that can be recompiled to deploy on multiple hardware platforms. Specifically, the robot side leverages ROS to take advantage of the built-in APIs provided by robot manufacturers~\cite{kinova}. We provide a detailed illustration of our system design in Fig.~\ref{fig:system_overview}(B). Processes on the same side communicate through a shared memory region, minimizing the overhead of disk I/O while maintaining process isolation and safety. Within each module, functionalities are instantiated using different algorithms and executed as individual threads. 
As shown in Fig.~\ref{fig:system_overview}(B) bottom, both XR and robot can be deployed on PCs or embedded devices, with at least one GPU equipped by XR device.
For clarity, we illustrate our system design using a stationary robot manipulator that has seven degree-of-freedom movement, though the framework is applicable to any robot with a similar joint configuration. For example, in the case of a humanoid robot, the user's fist gesture can be replaced with landmark positions, prompting the robot's hand to move accordingly.

\textbf{Network Communication.} XR-to-robot communication utilizes Data Distributed Services (DDS) for its compatibility with ROS and low transmission overhead when transmitting small payloads. DDS is preferable in XR-to-robot communication since XR only sends small payloads composed of user hand poses and fist gestures to remote robot. In contrast, robot-to-XR communication must handle significantly larger payloads including end-effector poses, gripper intensity, and point cloud data. We therefore adopt a UDP server-client design to avoid serialization in DDS, which incurs overwhelming overhead when transmitting large payloads~\cite{wang2019tzc}.

\subsection{Control Pipeline}

\textbf{XR Tracking.} Starting from the top left of Fig.~\ref{fig:system_overview}(C), TeleXR designs a \textit{hand tracking} to generate both a seven degree-of-freedom hand pose and a single fist score representing the fist gesture based on the Hand CAM image. The image is timestamped to signal the beginning of M2M latency and denoted as $t_{start}$. Hand tracking uses the Google Mediapipe~\cite{lugaresi2019mediapipe} model for its low memory requirements, but can be replaced with alternative algorithms and compatible Hand CAMs. Triggered by the completion of hand tracking, an \textit{XR talker} transmits the Hand CAM timestamp, user hand pose, and fist gesture to the remote robot, while hand tracking simultaneously publishes this information locally for XR visualization. A visual-inertial odometer (\textit{VIO}) and an IMU integration (\textit{IMUi}) in the same module are employed to generate user head pose for calculating viewport in XR visualization. Triggered by each Head CAM sensing, VIO uses OpenVINS~\cite{openvins} to generate a raw pose which is extrapolated by an IMUi running with GTSAM~\cite{gtsam} algorithm to produce head pose at the period of IMU. This extrapolation is necessary because users notice head motion misalignment within 20ms~\cite{illixr,atw}.

\textbf{Robot Control.} Following up to the top right of Fig.~\ref{fig:system_overview}(C), a \textit{robot listener}, running at the same period as the XR talker, receives all information from XR. It directly maps the user hand pose to the robot coordinate system and calculates its inverse kinematics to generate a waypoint. The robot listener locally maintains a \textit{target queue} that stores all received but not yet reached waypoints and fist gestures ordered by their $t_{start}$ to maintain correct pose sequences despite random order arrivals caused by network fluctuations. Before the start of teleoperation, the queue can buffer some initial information to prevent robot stops due to delayed information.

% \Johnson{Previous version of user motion reconstruction}As shown in \S\ref{subsec:Software_Architecture}, relying solely on the transmitted information make the system suffer greatly from any network dynamics. To battle against the problem, TeleXR designs a \textit{user motion reconstruction} that uses a trajectory planner algorithm to generate additional waypoints and determine the unique trajectory for the current robot manipulator motion given the current robot pose, velocity, and acceleration. It is scheduled to run at the same period as hand tracking in XR, as it is designed to make planning according to each hand pose and fist gesture. User motion reconstruction plans a trajectory with the received hand poses and generates the missing and delayed ones using the latest end-effector pose and velocity, which are given by the robot localization module. Upon determining the unique trajectory, it compiles a list of actuator control commands to achieve the trajectory and remove poses and gripper intensity that are reached from the queue published by robot listener. TeleXR implements the pure pursuit algorithm as the user motion reconstruction for its lightweight computation cost, but it can be replaced by other models as shown in Tab.~\ref{tab:functionalities}.\Hyoseung{Reconstruction should be the main item to discuss in this subsection, not as an extra piece of information. } 

The core functionality throughout the control pipeline lies in the \textit{user motion reconstruction} (pink box in Fig.~\ref{fig:system_overview}(C)), which reconstructs user motion on the robot side to mitigate the network problems highlighted in \S\ref{sec:Motivation}. User motion reconstruction is scheduled with the same period as hand tracking from XR, designed to process each generated hand pose and fist gesture. 
After getting the current end-effector pose, speed, and direction from robot localization, user motion reconstruction uses an online trajectory planner to generate the fine-grained waypoints between the current pose and the first waypoint in the target queue. Consequently, all the delayed and missing hand poses are compensated by the generated waypoints, so the original user motion is reconstructed even under network degradation. For gripper control, a linear intensity generator can be applied as the gripper is controlled by a single intensity value that can be easily extrapolated based on the latest gripper intensity.
Simultaneously, as the module generates waypoints to plan the trajectory, it issues control commands to the actuators to adjust the end-effector speed, direction, and gripper intensity.

\subsection{Feedback Pipeline}\label{subsec:Feedback_Pipeline}

% \textbf{Robot Localization.} Starting from the bottom right of Fig.~\ref{fig:system_overview}(C), We want to generate a pose that can represent the end-effector of the robot since we are only visualizing the end-effector and gripper in the frame output. Also, we want to obtain velocity information for the user motion reconstruction to use as input. As we want to maintain the freshness of the pose, we need to generate this information at the same rate as the maximum encoder frequency. Inside the localization, we designed a \textit{forward kinematics} functionality to calculate the end-effector pose based on the D-H parameter obtained from the robot manufacturer, and then directly sample the velocity using the encoder inertial data. For the gripper intensity, it also gets the gripper intensity from the encoder. At the end of forward kinematics, it compares the output pose and gripper with the queue published by robot talker, and removes the elements that are reached with an error bound of one cm based on manipulator inertia. Triggered by the forward kinematics, a \textit{robot talker} is applied for transmitting the end-effector pose and gripper intensity to XR. Meanwhile, it also reads from the robot CAM point cloud and transmits it if a new point cloud is generated, as the robot talker is running at a much smaller period than the robot CAM.

\textbf{Robot Localization.} Starting from Fig.~\ref{fig:system_overview}(C) bottom right, \textit{forward kinematics}~\cite{denavit1955kinematic} calculates the end-effector pose based on the robot manipulator's D-H parameters at the same period as the encoder. Meanwhile, the gripper intensity is directly sampled from encoder readings. Upon obtaining this information, the system removes the reached target waypoint and gripper intensity from the robot-side target queue using a designed error bound to prevent end-effector overshooting. Triggered by the forward kinematics, the \textit{robot talker} transmits the end-effector pose, gripper intensity, the latest point cloud scan from robot CAM, and the latest $t_{start}$ corresponding to the most recent user pose in the target queue to XR. To conserve bandwidth, the robot talker only transmits point cloud data at the period of Robot CAM.

\textbf{XR Visualization.} As illustrated in Fig.~\ref{fig:system_overview}(C) bottom left, XR visualization adopts an \textit{XR listener} that runs at the same period as the robot talker to receive all information from the robot. Central to the feedback pipeline is the \textit{robot pose reconstruction} functionality (red box in Fig.~\ref{fig:system_overview}(C)), which implements the second half of the dual-reconstruction paradigm to decouple XR visualization from network dependency.  

The robot pose reconstruction is performed in two steps. First, it extrapolates the latest received end-effector pose and gripper intensity to reconstruct an estimated pose and intensity based on the locally maintained user hand pose and fist gesture. Because the user information is more current compared to the received robot data and unaffected by network degradation, it provides a reliable basis for estimation over short time horizons. To determine the proper estimation window, robot pose reconstruction calculates the time difference between the $t_{start}$ associated with the latest received end-effector pose and the $t_{start}$ corresponding to the latest user hand pose. Currently, our implementation adopts an Extended Kalman Filter (EKF)~\cite{lee1994extended} algorithm for its lightweight computation, though a neural network prediction model~\cite{taud2017multilayer} could be used for highly randomized user motion. 

In the second step, robot pose reconstruction updates its model upon receiving new end-effector pose and gripper intensity. The model is adjusted based on the error between the received and estimated pose and intensity. In our EKF implementation, this corresponds to the update phase of the algorithm. When the prediction error becomes too large due to unforeseen external factors, the robot pose reconstruction pauses the next iteration to avoid abrupt visual changes. After the two steps, robot pose reconstruction outputs the estimated end-effector pose, gripper intensity and calculates the viewport based on user hand pose.

% The completion of robot pose reconstruction activates \textit{rendering} that generates a 2D frame using GPU. As humans are highly perceptive to motion misalignment, an additional step to reproject the 2D frame to produce a 3D frame that reflects the up-to-date user head pose is required. Since the head pose is constantly updated with IMUi that runs at a much smaller period, \textit{warpping} is used to calculate the view matrix based on the latest head pose at a period of 16ms. Upon completion of warpping, \textit{reprojection} reprojects the 2D frame based on the calculated view matrix to complete the 3D frame output.

The completion of robot pose reconstruction activates \textit{rendering}, which generates a 2D frame using GPU. As humans are highly perceptive to head motion misalignment, \textit{warpping} is used to calculate the view matrix based on the latest head pose at screen refresh rate, while \textit{reprojection} reprojects the 2D frame based on the calculated view matrix to complete the 3D frame output.

% \begin{figure}[t]
% \includegraphics[width=\linewidth]{figs/gap_filling.pdf}
% \caption{TODO}
% \label{fig:gap_filling}
% \end{figure}

\subsection{TeleXR System Optimization}\label{subsec:TREX2_System_Optimization}

\begin{figure}[t]
\includegraphics[width=\linewidth]{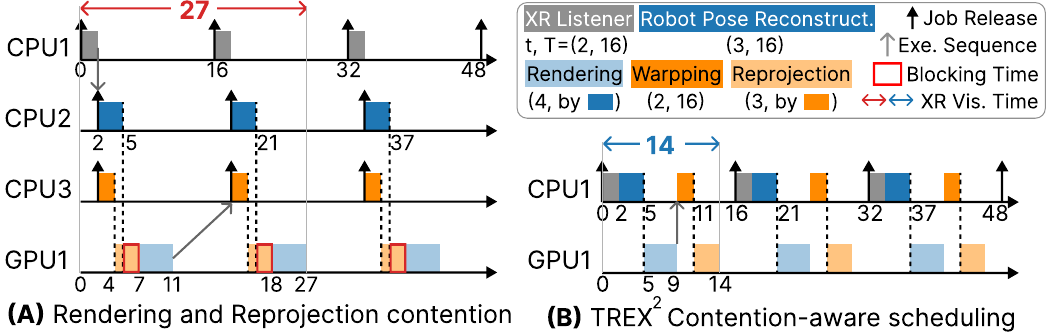}
\caption{Contention-aware scheduling example.}
\label{fig:system_optimization}
\end{figure}

% \textbf{Contention-aware scheduling.} As shown in \S\ref{subsec:Network_and_Hardware_Challenges}, hardware contention exists when multiple functionalities within a module compete for one computation resource like a GPU stream. To avoid this contention, TeleXR adopts a contention-aware scheduling which synchronize the execution of the functionalities that are competing for the same GPU stream. The contention-aware scheduling first identifies the contention functionalities with GPU traces software and record the execution time and periods of each functionalities denoted as $\{t_1, t_2,...,t_n\}$ and $\{T_1, T_2,...,T_n\}$. Following the execution order shown in Fig.~\ref{fig:system_overview}, it finds the first functionality with period $T_1$. It schedules the first job of $t_1$ to follow the completion of its upstream functionality. Then, we start execute the entire sequence of $t_2$ to $t_n$ in a single thread to prevent any out-of-order execution that breaks the pipelines. Lasty, it sets a lower bound of $T_1\geq\sum_{i=1}^{n-1}t_i$ to eliminate the possibility of choosing a too small period that makes the entire sequence unable to complete.

\textbf{Contention-aware Scheduling.} 
To address the hardware contention issue discussed in \S\ref{subsec:Network_and_Hardware_Challenges}, TeleXR adopts the contention-aware scheduling approach~\cite{zhang2024boxr} to synchronize functionalities that compete for the GPU. During the setup phase, if potential contention is detected in a module, the scheduler identifies all $n$ functionalities within the same module process and orders them by their execution sequence as shown in Fig.~\ref{fig:system_overview}(C). The scheduler then profiles the functionalities $\{\tau_1,\tau_2...,\tau_n\}$, recording their independent execution times as $\{t_1, t_2,...,t_n\}$ and target periods $\{T_1, T_2,...,T_n\}$. For execution time, 99\%-percentile observed values are recorded to maintain scheduling valid under runtime dynamics. 

The scheduler establishes a lower bound on $T_1$ such that $T_1\geq\sum_{i=1}^{n-1}t_i$ to prevent selecting a period that is too short for all $n$ functionalities to complete. Subsequently, the scheduler triggers $\tau_2$ upon completion of $\tau_1$, $\tau_3$ upon completion of $\tau_2$, and so on until $\tau_n$ is scheduled. This sequential activation prevents execution order inversion, ensuring that downstream functionalities process up-to-date information to reduce M2M latency.

Fig.\ref{fig:system_optimization}(A) illustrates an example in XR visualization where rendering and reprojection compete for the same GPU stream by writing to the same memory region of the frame. This contention causes blocking time and outdated frames, which result in a longer XR visualization runtime of 27 units. As shown in Fig.\ref{fig:system_optimization}(B), the contention-aware scheduler synchronizes the execution of all functionalities in XR visualization by enforcing the correct sequence and triggering each downstream functionality only after its predecessor completes. It also verifies that all functionalities finish execution within one XR listener period. With this approach, the XR visualization runtime is reduced to 14, reducing the time to generate a 3D frame by 7 units compared to Fig.~\ref{fig:system_optimization}(A).

\textbf{Bandwidth-adaptive Point Cloud Scaling.} As described in \S\ref{subsec:Network_and_Bandwidth Dynamics}, a solution is required to reduce the point cloud size while preserving visual quality for human perception during XR visualization. Because both environmental context and object features are necessary to convey spatial information around a remote robot, simply extracting and transmitting object features would result in the loss of environmental details. However, since human perception is particularly sensitive to the structural integrity of a point cloud~\cite{lu2022point}, maintaining edge points while reducing interior points can decrease the overall point cloud size without significantly degrading perceptual quality. Based on this insight, we propose a point cloud scaling algorithm that selectively reduces interior points using a dynamic scaling factor $r$.

During the setup phase, the available bandwidth $B$, the size of a single point $b$, and the robot talker period $T_{rt}$ are profiled and remain fixed at runtime. These parameters determine the maximum number of points that can be transmitted in a single point cloud as $N_{max}=\frac{B\,T_{rt}}{b}$. At runtime, upon scanning each point cloud from Robot CAM, TeleXR checks whether the number of points exceeds $N_{max}$. If it does, point cloud scaling is executed within the localization module at the start of the robot talker. The algorithm first applies Canny edge detection~\cite{canny1986computational} to the latest point cloud to categorize the points into edge points, denoted as $N_e$, and interior points, denoted as $N_{in}$. To satisfy the bandwidth constraint, the scaling factor $r$ is chosen to reduce only the interior points so that $N_e + r\,N_{in} \leq N_{max}$. Since the size of the point cloud and its number of points follow an inverse linear relationship, we define $r\in[1,0)$ as follows:
\begin{equation}
r = \min\left\{1, \frac{N_{{max}} - N_e}{N_{in}}\right\}\label{eq:r}
\end{equation}

\section{Conclusion}

Motivated by the degradation in teleoperation quality caused by strong network dependency as discussed in \S\ref{sec:Motivation}, TeleXR adopts a novel approach by reconstructing counterpart information on both XR and robot sides. This enables the system to overlap the execution of downstream modules with network latency as detailed in \S\ref{subsec:Software_Architecture} to \S\ref{subsec:Feedback_Pipeline}. Moreover, TeleXR optimizes system performance through contention-aware scheduling and point cloud scaling as described in \S\ref{subsec:TREX2_System_Optimization}, which mitigate hardware contention and overcome bandwidth limitations.

\bibliography{ref}
\bibliographystyle{IEEEtran}

%%%%%%%%%%%%%%%%%%%%%%%%%%%%%%%%%%%%%%%%%%%%%%%%%%%%%%%%%%%%%%%%%%%%%%%%%%%%%%%%
\end{document}